# APPLYING MULTI-ANGLED PARALLELISM TO SPANISH TOPOGRAPHICAL MAPS

*Josep-Maria Cuscó, Marcos Faúndez*
Departament de Teoria del Senyal i Comunicacions
ETSE Telecomunicació (Universitat Politècnica de Catalunya)
c/ Gran Capità, s/n, E08034 Barcelona, Spain
e-mail: marcos@gps.tsc.upc.es

## ABSTRACT

Multi-Angled Parallelism (MAP) is a method to recognize lines in binary images. It is suitable to be implemented in parallel processing and image processing hardware. The binary image is transformed into directional planes, upon which, directional operators of erosion-dilation are iteratively applyed. From a set of basic operators, more complex ones are created, which let to extract the several types of lines. Each type is extracted with a different set of operations and so the lines are identified when extracted.
In this paper, an overview of MAP is made, and it is adapted to line recognition in Spanish topographical maps, with the double purpose of testing the method in a real case and studying the process of adapting it to a custom application.

## 1 INTRODUCTION

Lines are elements of arbitrary length and direction. Their position is the whole set of points along their path, and their features are the thickness, shape and color of their stroke.

Line recognition is usually made upon vectorized images ([Kasturi 92]), but in the vectorization process the two-dimensional information of the image is lost ([Yamada 93]). In map and drawing processing, this information is important to know the spatial relationship among the elements.

There are some methods which extract lines directly from raster images: local operators (filters), Hough transform ([Hough 62]), etc. They perform well with straight lines but their great directionality prevents them from extracting lines which take several directions along their path. Moreover they don't give the features of lines (they extract lines but they don't identify them).

## 2 THE METHOD

MAP (Multi-Angled Parallelism) is a method developed by H.Yamada and K.Yamamoto (see [Yamada 91] and [Yamada 93]). The authors define it as an unification of the image representation by directional planes and non isotropic neighborhood morphological operations. MAP operates upon binary images and it is conceived as a highly parallel method, suitable to parallel computing and hardware embedding, which enables high speed line recognition.

The binary image must be converted to an arrangement in which every pixel carries information of several directions, one direction for each bit. The set of bits of each direction make a directional plane, as figure 1a shows.

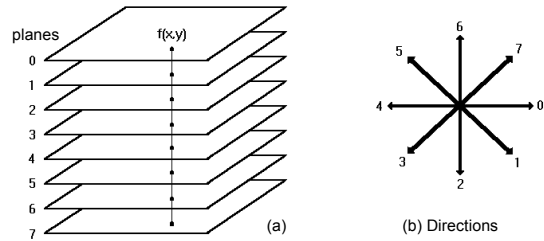

*Figure 1. Image representation by directional planes.*

In map processing, eight directions are enough. This is also the most convenient number from a practical point of view, since in the rectangular grid each pixel has eight immediate neighbors and eight bits form a byte.

Directional planes are obtained as the contours of a binary image in the given directions:

$$f_d = b \cap \overline{b[d+4]}$$

$f_d$ : plane of direction d (see figure 1a).
d : direction from 0 to 7 (see figure 1b).
b : binary image.
f[d] : neighborhood of f in direction d, obtained by shifting the image one pixel in direction d+4.
b[d+4] means shifting b one pixel in direction d (module 8).

Figure 2 shows the directional planes resulting from applying this formula to an example binary image.

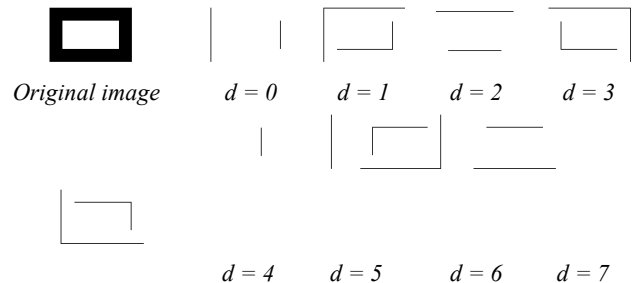

*Figure 2. Transforming an image into directional planes. (black=1, white=0)*

Upon the directional planes, directional operators are applied. Directional operators are erosion and dilation local operators, applied on non-isotropic neighborhoods, oriented according to the direction of the plane. To get a good resolution and a low computing charge, small neighborhoods are used and operators are applied iteratively. In figure 3 the basic set of erosion-dilation operators is presented with their respective neighborhoods in direction 0.

Nondirectional operators (n)
(4 (a) and 8 (b) - neighbors ) :
$$D_{n4}f = f \cup (\bigcup_{d=0,2,4,6} f[d])$$
$$D_{n8}f = f \cup (\bigcup_{d=0}^{7} f[d])$$
$$E_{n4}f = f \cap (\bigcap_{d=0,2,4,6} f[d])$$
$$E_{n8}f = f \cap (\bigcap_{d=0}^{7} f[d])$$

Fan directional operators (>d)
(neighborhood (d) ) :
$$D_{>d}f = f \cup (\bigcup_{i=d+3}^{d+5} f[i])$$
$$E_{>d}f = f \cap (\bigcup_{i=d-1}^{d+1} f[i])$$

Single directional operators
(=d) (neighborhood (c) ) :
$$D_{=d}f = f \cup f[d+4]$$
$$E_{=d}f = f \cap f[d]$$

Directional operators orthogonal to d (=*d, >*d)
(neighborhoods (e) y (f) ) :
$$D_{=*d}f = D_{=d-2}f \cup D_{=d+2}f$$
$$D_{>*d}f = D_{>d-2}f \cup D_{>d+2}f$$
$$E_{=*d}f = E_{=d-2}f \cap E_{=d+2}f$$
$$E_{>*d}f = E_{>d-2}f \cap E_{>d+2}f$$

These operators may be applied iteratively ...
$$D^2 f = D(Df)$$
$$E^3 f = E(E(Ef))$$

...and masked with image 'g':
$$D_{:g}f = g \cap Df$$
$$E_{:g}f = E(f \cup g)$$

Macro - operators open and close are also defined :
$$Open f = D(Ef)$$
$$Close f = E(Df)$$

that may be also masked:
$$Open_{:g}f = D_{:g}(Ef)$$
$$Close_{:g}f = E(D_{:g}f)$$

... with the source image indeed:
$$Open_{:*}f = D_{:f}(Ef)$$

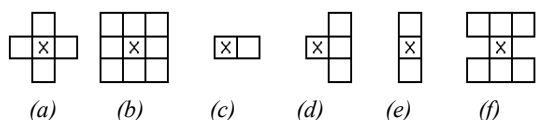

*Figure 3. Set of erosion-dilation operators.*

Note that Minkowski subtraction (intersection of shiftings) is used instead of erosion because it can be quickly computed, specially with image processing hardware. However, the fan directional erosion is not consistent since it is defined as the intersection of the image with the union of the shiftings. The authors justify the change because of the need for propagation of the ends of lines. But this modification breaks the duality erosion-dilation, affecting open and close operations: with the real erosion, the image resulting from an open is contained in the original, but with the modification this no longer holds.

To prevent uncontrolled line propagation, masked open and close operations are used.

## 3 LINE EXTRACTION

MAP line extraction is based on the principles of *straightness* and *connectedness*. Straightness is the assumption that in important sections, the line doesn't change of direction in a significant way. Straightness allows to consider the line as a set of one-directional segments. These segments are extracted from each directional plane with directional operations, which are combinations of the directional operators shown in figure3. Connectedness means that there is a contact among the points in a line and therefore among segments of different directions. It allows to connect the segments of a line that have been extracted in each directional plane. It also helps to extract sharp curves.

Given a binary image, the first step is its conversion into directional planes :
$$f_d = b \cap \overline{b[d+4]}$$

Directional planes contain mainly lines of direction perpendicular to the direction of the plane, but they also contain lines of neighbor directions (see figure 2). Directionality is enforced with the Edge operation, which extracts cores of segments (sections with strict directionality) :
$$f_d^{edge} = f_d \cap (f_d[d-2] \cup f_d[d+2])$$

The Edge operation is greatly affected by noise pixels along the contour, which remove the straightness and cause interruptions in the segments extracted by Edge operation. Short operation tries to re-join broken segments, expanding them along the segments extracted in the directional planes. Observe that the Open operation is masked with the same image upon which it is applied, to prevent uncontrolled line propagation.

$$f_d^{short} = Open_{>*d:*}^2 \left\{ f_d^{edge} \cup E_{>*d}^2 \begin{pmatrix} f_d^{edge} \cup \\ D_{>d-2:f_d}^2 End_{d-2} f_d^{edge} \cup \\ D_{>d+2:f_d}^2 End_{d+2} f_d^{edge} \end{pmatrix} \right\}$$

$$End_d f = f \cap \overline{E_{>d}f}$$

Some segments extracted with the Short operation may be part of a larger segment which has been interrupted in the directional plane because of a sudden change of direction or a line intersection. To remove those interruptions, the same procedure as Short is applied but masking dilations with the original binary image to allow the closing of such interruptions.

$$f_d^{middle} = Open_{>*d:*}^3 \left\{ f_d^{short} \cup E_{>*d}^2 \begin{pmatrix} f_d^{short} \cup \\ D_{=d-2:b} D_{>d-2:b} End_{d-2} f_d^{short} \cup \\ D_{=d+2:b} D_{>d+2:b} End_{d+2} f_d^{short} \end{pmatrix} \right\}$$

Middle is the basis to extract the various types of lines:
- Solid lines: segments which surpass a given length are extracted:

$$f_d^{longb} = Open_{>*d:*}^{12} f_d^{middle}$$

- Stippled lines: their segments must be connected. This is got by allowing the free expansion of segments:

$$f_d^{longw} = Open_{>*d:*}^{12} \left\{ f_d^{middle} \cup E_{>*d}^4 \begin{pmatrix} f_d^{middle} \cup \\ D_{=d-2}^3 D_{>d-2} End_{d-2} f_d^{middle} \cup \\ D_{=d+2}^3 D_{>d+2} End_{d+2} f_d^{middle} \end{pmatrix} \right\}$$

By the principle of connectedness, the segments of a given class extracted in the directional planes are joined to get the whole line:

$$f = \bigcup_{d=0}^{7} f_d$$

## 4 LINE RECOGNITION

Line recognition is achieved by selective extraction of the lines in an image. Combining the operations that we have already presented, more complex operations are devised, suitable to extract each type of line.

To illustrate the method, we are going to recognize lines in a real image. We have chosen a topographical map because it is a sort of document with a lot of meaningful lines of different length and features. The *Mapa Topográfico Nacional de España* is a set of spanish topographical maps edited by the *Instituto Geográfico Nacional* in 1:25.000 scale. This set of maps covers almost all Spain and it is specially attractive because its format is highly consistent across all maps. This means that operations developed to recognize lines in one map are directly applicable to the whole set.

We scan the map at 300dpi. To overcome the limitations of our scanning device, preprocess is necessary, including an alignment of the RGB components (see [Cuscó 95]). Next, the image is converted to HSI color space for color separation. Elements in the map are placed in different images according to their color. This is a first segmentation of the image, based on the color feature.

Now, MAP can be applied to these binary images containing lines of a certain color. As an example we take a piece (250x250 pixels) of the image with black elements. In it, there are two types of linear elements (see figure 4): a track (represented with a double line solid-stippled) and several paths (symbolized with stippled lines). We apply the development of section 3 to this image, extracting solid and stippled lines (figure 5 shows the process in direction 1, figure 6 the process in direction 3 and figures 7a and 7b the solid and stippled lines extracted). From this primitives, we have created special operations for extracting each type of linear element:
- As tracks are represented with two parallel lines (solid-stippled), their route is the space between both:

$$f^{stippled} = Close_{n8}^2 f^{longw}$$
$$f^{total} = f^{longb} \cup f^{stippled}$$
$$f^{closed} = Close_{n8}^2 f^{total}$$
$$f^{track} = D_{n8:f^{closed}}^2 (f^{closed} \cap \overline{f^{total}})$$

- Paths are all the stippled lines that are not part of a track:

$$f^{path} = f^{stippled} \cap \overline{D_{n8}^3 f^{track}}$$

The results are shown figures 7c and 7d. The track has been recognized correctly but in the paths there are some interruptions caused by segment loss. There is also some wrong recognition due to non-linear elements (characters) of size similar to the segments of stippled lines.

Operators in sections 2 and 3 come from [Yamada 93], with the coefficients adapted to our application.

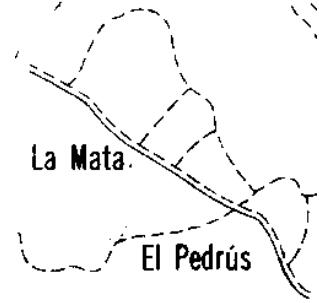

*Figure 4.* *Binary image of 256x256 pixels.*

## 5 CONCLUSIONS

The main advantage of MAP is that it is a low-level method which, implemented with appropriate software and hardware, can perform a high speed line recognition. If the operations are implemented as the formulae (plane shifting), recognition time is independent of the number of lines in the image and the size of the image (provided that it fits in the frame buffer).

MAP is good at extracting large sets of solid lines with slow change in direction, but it appears weak when its two principles don't hold: step curves (absence of straightness), stippled lines (no connectedness) or worse: both at a time.

Since the method works with contours, it is seriously affected by noise and line intersection.

These drawbacks keep it from become a reliable stand-alone line recognition method, but it can be a good heavy duty component in a recognition system. It should be necessary a preprocess to remove contour noise and non linear elements, and to thin lines. Then, MAP would perform the mechanical line recognition, and finally, a high-level method should interpret the raw information to correct wrong assignations.

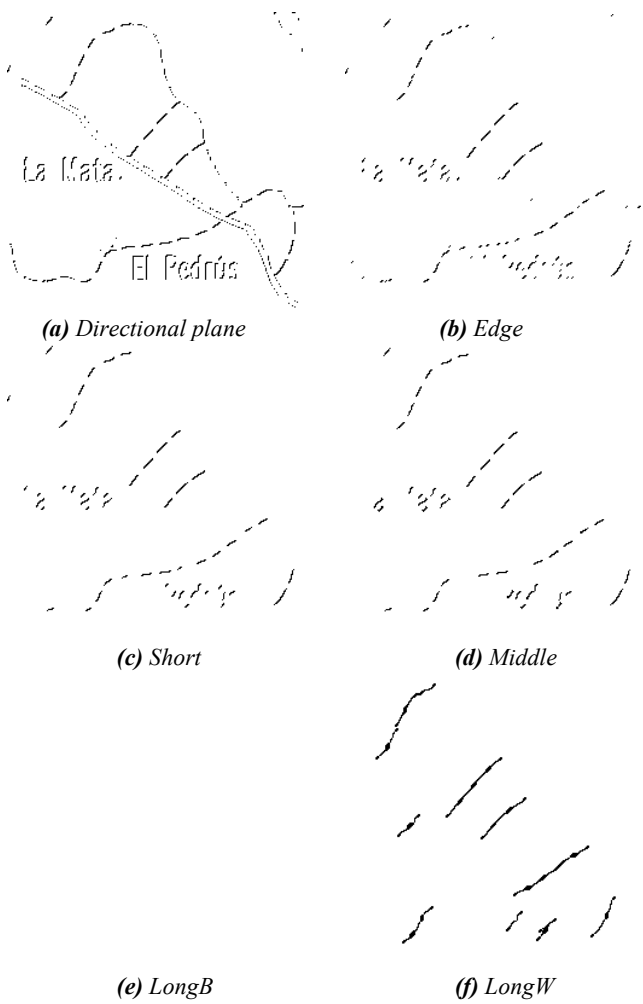

*Figure 5. Line extraction in direcional plane 1.*

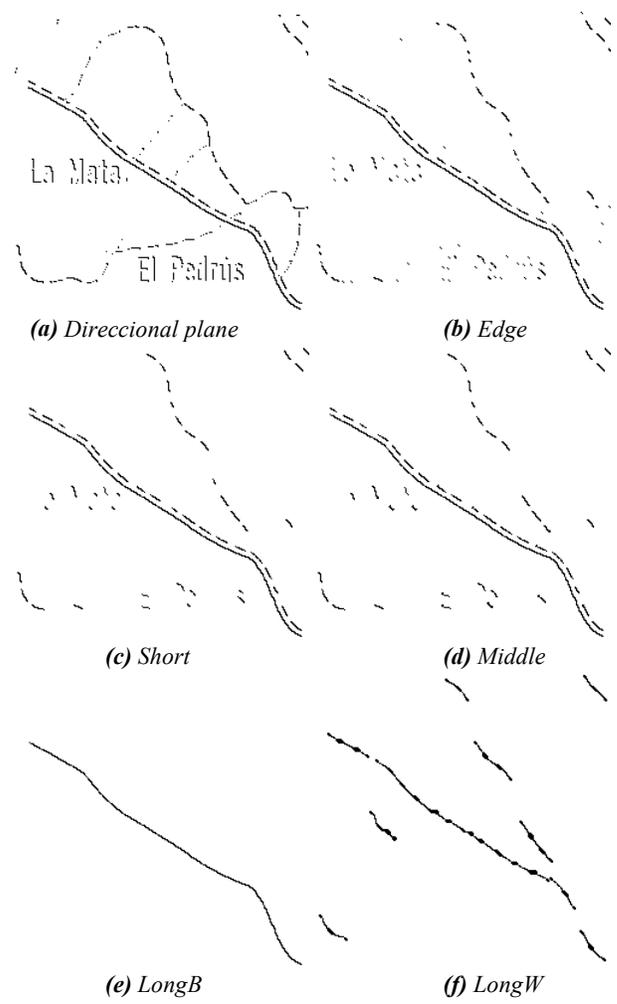

*Figure 6. Line extraction in direcional plane 3.*

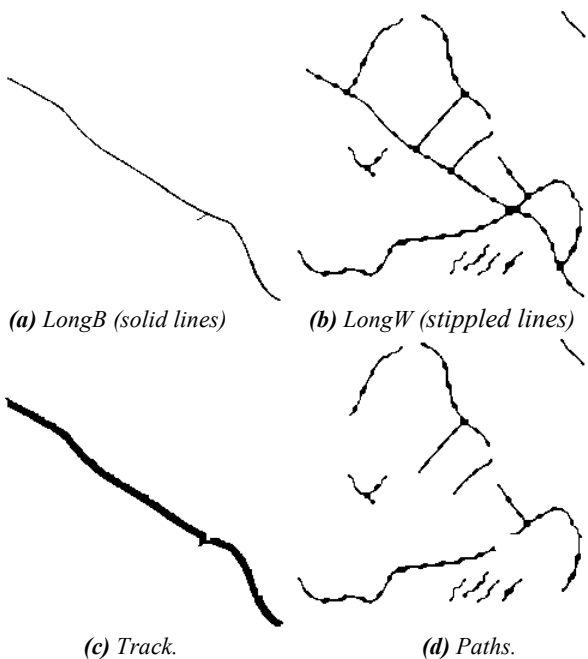

*Figure 7. Recognition results.*